\pdfoutput=1
\documentclass[11pt]{article}

\usepackage[final]{acl}
\usepackage{tablefootnote}
\usepackage{times}
\usepackage{latexsym}
\usepackage{threeparttable}
\usepackage{amsmath}
\usepackage{multirow}
\usepackage{booktabs}
\usepackage{stfloats}
\usepackage[T1]{fontenc}
\usepackage{soul} 

\usepackage{color, xcolor}
\usepackage[utf8]{inputenc}
\usepackage{enumitem}
\setenumerate[1]{itemsep=1.5pt,partopsep=0pt,parsep=\parskip,topsep=5pt}
\setitemize[1]{itemsep=1.5pt,partopsep=0pt,parsep=\parskip,topsep=5pt}
\usepackage{microtype}

\usepackage{inconsolata}

\usepackage{graphicx}
\title{BioHopR: A Benchmark for Multi-Hop, Multi-Answer Reasoning in Biomedicine}

\author{
    \textbf{Yunsoo Kim}$^{1}$ \quad
    \textbf{Yusuf Abdulle}$^{1,2}$ \quad
    \textbf{Honghan Wu}$^{1,3}$ \\
    $^1$UCL \quad
    $^2$King's College London \quad
    $^3$University of Glasgow \\
    \texttt{\{yunsoo.kim.23, honghan.wu\}@ucl.ac.uk}
}
\begin{document}
\maketitle
\begin{abstract}
Biomedical reasoning often requires traversing interconnected relationships across entities such as drugs, diseases, and proteins. Despite the increasing prominence of large language models (LLMs), existing benchmarks lack the ability to evaluate multi-hop reasoning in the biomedical domain, particularly for queries involving \textit{one-to-many} and \textit{many-to-many} relationships. This gap leaves the critical challenges of biomedical multi-hop reasoning underexplored. To address this, we introduce \textbf{BioHopR}, a novel benchmark designed to evaluate multi-hop, multi-answer reasoning in structured biomedical knowledge graphs. Built from the comprehensive PrimeKG, BioHopR includes 1-hop and 2-hop reasoning tasks that reflect real-world biomedical complexities. 

Evaluations of state-of-the-art models reveal that O3-mini, a proprietary reasoning-focused model, achieves \textbf{37.93\%} precision on 1-hop tasks and \textbf{14.57\%} on 2-hop tasks, outperforming proprietary models such as GPT4O and open-source biomedical models including HuatuoGPT-o1-70B and Llama-3.3-70B. However, all models exhibit poor capabilities in the multi-hop reasoning, underscoring the challenges of resolving implicit reasoning steps in the biomedical domain. By addressing the lack of benchmarks for multi-hop reasoning in biomedical domain, BioHopR sets a new standard for evaluating reasoning capabilities and highlights critical gaps between proprietary and open-source models while paving the way for future advancements in biomedical LLMs. BioHopR is available at \url{https://huggingface.co/datasets/knowlab-research/BioHopR}.

\end{abstract}

\section{Introduction}
\label{sec:intro}

Recent advances in large language models (LLMs) and Question Answering (QA) systems have shifted the focus from simple factoid retrieval tasks to more sophisticated reasoning capabilities \cite{huang2022towards,plaat2024reasoning,openai_o3_mini}. Among these, \textbf{multi-hop reasoning} has emerged as a critical area of research, where answering a question requires traversing multiple interconnected reasoning steps \cite{misra2023triggering,yang2024latent,schnitzler2024morehopqa}. For example, to answer \textit{“Who is the wife of the president of the United States?”}, a LLM must first identify the president (step 1) and then determine their spouse (step 2). This type of reasoning, referred to as multi-hop reasoning, is especially vital in domains where information is highly interconnected, such as the biomedical field.

In the biomedical domain, knowledge is often structured in ontologies and knowledge graphs (KGs), where entities like drugs, diseases, proteins, and phenotypes are represented as nodes, and their relationships as edges \cite{himmelstein2017hetionet,sung2021can,chandak2023primekg}. Biomedical queries frequently demand multi-step reasoning over these graphs \cite{sung2021can,su2024knowledge,matsumoto2025escargot}. For instance, identifying diseases associated with a drug might require a single-hop relation, while determining proteins targeted by that drug through its associated disease involves two reasoning steps. Furthermore, biomedical reasoning often involves one-to-many or many-to-many relationships. For example, a single question might have several correct answers - like a drug that works on multiple proteins \cite{liang2019predicting}. This complexity highlights the need for specialized benchmarks that rigorously evaluate models' ability to reason across multiple steps while generating comprehensive, multi-answer responses.

Existing benchmarks for multi-hop reasoning, such as Hetionet \cite{himmelstein2017hetionet} and other biomedical QA datasets \cite{biomedicalmultihop}, have laid the groundwork for evaluating multi-hop capabilities in the biomedical domain. However, these benchmarks primarily focus on single-hop tasks or utilize pre-defined templates that fail to fully capture the intricacies of multi-step reasoning. Similarly, general-domain benchmarks like TWOHOPFACT \cite{yang2024latent} test models' latent multi-hop reasoning ability but lack the domain-specific challenges of biomedical reasoning, such as reasoning over structured relationships and handling multi-answer outputs. As a result, the unique challenges of biomedical multi-hop reasoning remain underexplored.

To address these limitations, we introduce \textbf{BioHopR}, a new benchmark specifically designed to test the multi-hop reasoning capabilities of LLMs in the biomedical domain. Unlike general-domain benchmarks that rely on reasoning across disconnected documents, BioHopR focuses on reasoning within a single, structured biomedical knowledge graph. Our benchmark systematically constructs 1-hop (e.g., Drug–Disease) and 2-hop (e.g., Drug–Disease–Protein) question-answer pairs from the PrimeKG knowledge graph \cite{chandak2023primekg}. Questions are designed to evaluate models' abilities to reason step-by-step, explicitly requiring the inference of intermediate entities, and generate multi-answer responses reflective of real-world biomedical complexity.

\paragraph{Contributions.}
Our main contributions are as follows:
\begin{itemize}[noitemsep, topsep=0pt]
    \item \textbf{A New Benchmark for Multi-Hop Reasoning:} We propose \textbf{BioHopR}, the first publicly available benchmark explicitly designed to evaluate multi-hop, multi-answer reasoning within structured biomedical knowledge graphs. We will release the dataset and the code for evaluation.
    \item \textbf{Multi-hop Knowledge Curated from a Comprehensive Up-To-Date Knowledge Graph:} Leveraging the comprehensive and up-to-date PrimeKG knowledge graph, we systematically construct a dataset of 1-hop and 2-hop biomedical questions and their answers, ensuring real-world relevance.
    \item \textbf{Evaluation and Analysis of LLMs in Biomedical Multi-hop Reasoning:} We evaluate state-of-the-art LLMs on our benchmark, highlighting their strengths and, more importantly, limitations in handling biomedical multi-hop reasoning tasks.
\end{itemize}

By introducing BioHopR, we aim to fill a critical gap in multi-hop QA research and advance the development of LLMs capable of robust and interpretable reasoning in structured, high-stakes domains like biomedical research and healthcare.

\begin{table*}[htbp]
\centering
\begin{tabular}{p{5.8cm}p{1.6cm}p{4.6cm}p{2.3cm}}
\hline
\textbf{Dataset} & \textbf{Domain} & \textbf{Reasoning} & \textbf{Answer-Level} \\
\hline
MedQA \cite{jin2021disease}& Biomedical & No & Single Answer \\ 
Hetionet QA \cite{himmelstein2017hetionet} & Biomedical & Graph-based Reasoning (MH) & Single Answer \\
MedExQA \cite{kim2024medexqa} & Biomedical & Explanation Generation & Single Answer \\
TWOHOPFACT \cite{yang2024latent} & General & Implicit Reasoning (MH) & Single Answer \\
\textbf{BioHopR (Ours)} & Biomedical & Implicit Reasoning (MH) & Multi Answers \\
\hline
\end{tabular}
\caption{Comparison of BioHopR with existing datasets. Key differentiators include domain focus, reasoning type (MH is tagged for multi-hop reasoning supported dataset), and answer-level, such as multi-answer capability.}
\label{tab:dataset_comparison}
\end{table*}

\section{Related Works}
\label{sec:related_work}

\paragraph{Biomedical Question Answering.}
Research in medical LLMs has been facilitated by the development of question-answering (QA) datasets that benchmark models' understanding of medical domain knowledge \cite{hendrycks2020measuring, jin2021disease, pal2022medmcqa}. These datasets typically consist of multiple-choice questions (MCQs) focused on single-hop reasoning tasks, providing a straightforward way to evaluate LLMs' ability to comprehend and respond to diverse medical inquiries. While these benchmarks have driven significant progress, they primarily measure classification accuracy, which is insufficient for capturing the nuanced reasoning required for medical expertise.

Medical QA often involves interconnected concepts where reasoning over multiple steps is crucial. However, current benchmarks rarely go beyond single-hop tasks and do not evaluate models' ability to provide explanations for their answers or justify their reasoning process. Recently, MedExQA introduced an evaluation framework with detailed explanations for assessing the reasoning capabilities of LLMs \cite{kim2024medexqa}. While this is a step forward, it remains constrained to single-hop reasoning and does not address the need for multi-hop reasoning or the generation of multiple valid answers—a common requirement in biomedical inquiries.

\paragraph{Knowledge Graph Question Answering.}
Knowledge Graph Question Answering (KGQA) systems leverage structured knowledge graphs to answer questions that require reasoning over graph-based relationships. In the biomedical domain, Hetionet \cite{himmelstein2017hetionet} introduced a knowledge graph containing entities like genes, drugs, and diseases, enabling structured reasoning. Extensions of Hetionet have been used for multi-hop QA tasks \cite{biomedicalmultihop}, but these datasets often rely on fixed templates and predefined reasoning paths, limiting their ability to evaluate the nuanced multi-hop reasoning required in real-world biomedical applications.  This work explored techniques such as knowledge graph embeddings and graph neural networks \cite{kipf2016semi,hamilton2018embedding}, and transformer-based models like BioBERT \cite{lee2020biobert} to extract and utilize graph-based knowledge. However, this dataset tests the model's performance in a classification task to a single answer. Also, this dataset is not publicly available, limiting its role in facilitating biomedical large language model research. 

In domains like biomedical science, many questions inherently involve multiple correct answers. For instance, identifying all drugs that treat a specific disease or all proteins associated with a disease phenotype requires models to retrieve comprehensive sets of answers rather than a single response. 

\paragraph{Latent Multi-Hop Reasoning in Large Language Models.}
Recent work has explored the latent reasoning capabilities of LLMs, focusing on whether models can implicitly infer intermediate entities and use them for multi-step reasoning. The TWOHOPFACT dataset \cite{yang2024latent} evaluates this capability by testing whether LLMs can identify "bridge entities" in two-hop reasoning tasks. While TWOHOPFACT demonstrates that LLMs can perform latent multi-hop reasoning in general domains, it does not address the unique challenges of biomedical reasoning. Biomedical queries often require explicit reasoning over structured data and demand comprehensive answers involving one-to-many or many-to-many relationships.

These gaps highlight the need for a benchmark like BioHopR, which explicitly evaluates models' ability to perform step-by-step reasoning and generate multi-answer outputs in the biomedical domain.

\paragraph{Multi-Answer Reasoning.}
Existing QA benchmarks, both in general and biomedical domains, typically assume a one-to-one mapping between questions and answers, which oversimplifies the complexity of real-world reasoning tasks. This assumption is especially problematic in the biomedical domain, where relationships between entities are often one-to-many or many-to-many. 

\paragraph{BioHopR}
BioHopR addresses this limitation by introducing questions that require multi-answer reasoning, ensuring that the benchmark captures the intricate relational structures and knowledge dependencies present in biomedical science. The differences between our dataset and relevant datasets are summarized in Table \ref{tab:dataset_comparison}.

\section{BioHopR: Multi-hop Reasoning in Biomedicine}
\label{sec:benchmark}

\begin{figure}[htbp]
\centering
\includegraphics[width=\linewidth]{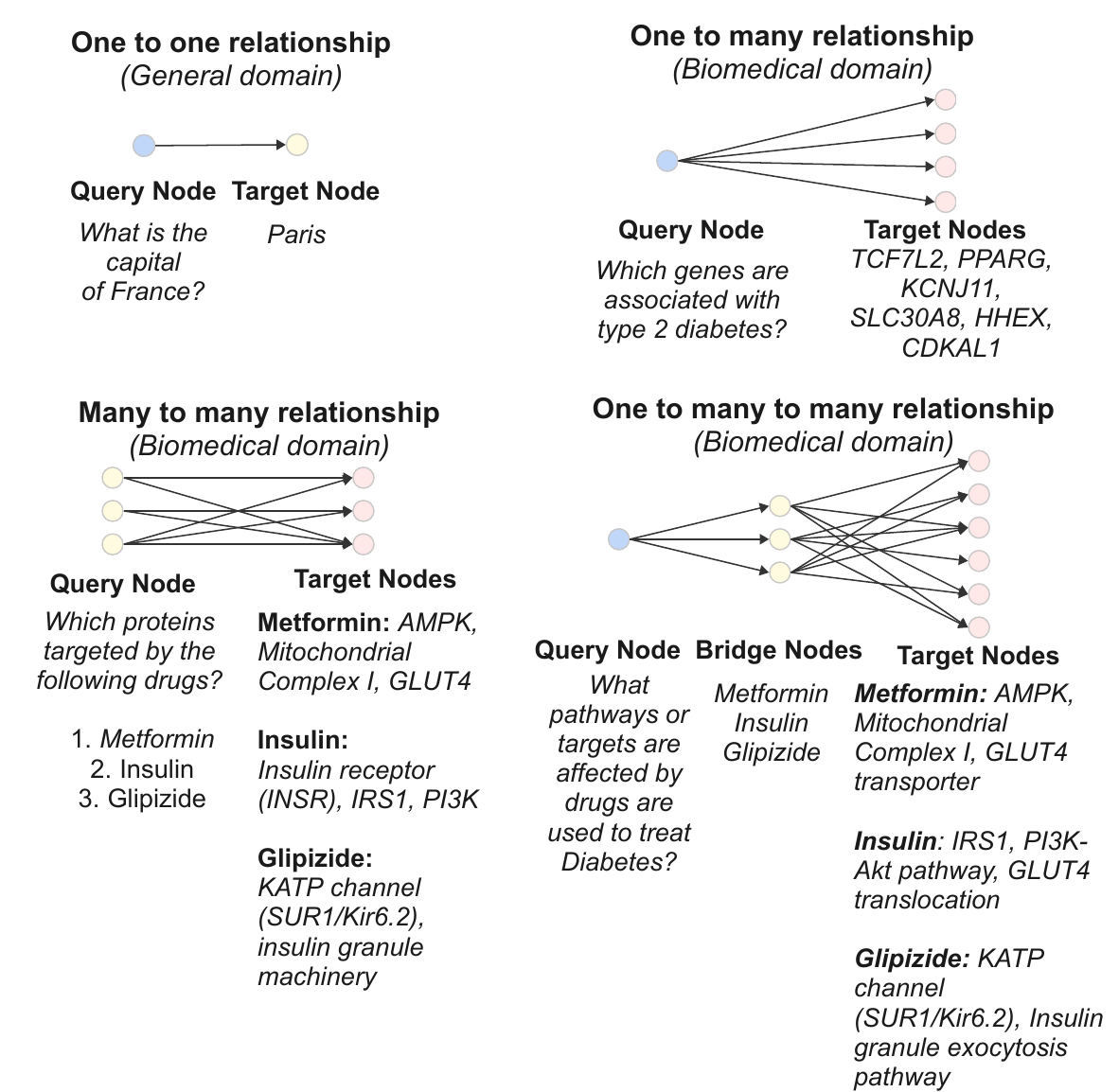}
\caption{Illustration of the relationships observed in this work using a biomedical knowledge graph. Each subplot visualizes a different type of question–answer relationship, highlighting the complexity and variability inherent in biomedical data.}
\label{fig:relationships}
\end{figure}

\textbf{BioHopR} is a benchmark specifically designed to evaluate the ability of large language models (LLMs) in performing multi-hop reasoning and generating multi-answer outputs in the biomedical domain. Compared to other knowledge graphs such as Hetionet, PrimeKG provides a broader coverage of biomedical entities, richer relational structures, and up-to-date knowledge in the field \cite{chandak2023primekg}. This allows for the generation of diverse, clinically relevant up-to-date multi-hop queries. By carefully creating questions over \textit{PrimeKG}, the dataset is designed to follow a \textbf{one-to-many-to-many} relationship structure, where one question can have several correct answers. This restriction ensures that queries reflect real-world biomedical scenarios where entities like drugs, diseases, and proteins exhibit hierarchical and complex interconnected relationships.

\subsection{Multi-hop, Multi-answer Knowledge Formalization}
\paragraph{Nodes and Relations.}
In our dataset, the entities in PrimeKG are represented as nodes, and their relationships are directed edges. For any query, the node from which reasoning starts is defined as the \textbf{query node}, and the node(s) forming the final answers are the \textbf{target nodes}. In the case of 2-hop reasoning, the intermediate node connecting the query and target is defined as the \textbf{bridge node}. We restrict node types to the following: Drug, Proteins, Disease, Phenotype.

\paragraph{Relationship Structure.}
As seen in Figure \ref{fig:relationships}, the dataset is restricted to follow a \textbf{one-to-many-to-many} relationship structure, where each question has  multiple correct answers and each answer may relate to several different entities.

In 1-hop questions, a direct relationship connects the \textbf{query node} to the \textbf{target nodes}. For example:
\begin{equation}
    \text{Query (Drug)} \xrightarrow{\text{treats}} \text{Target (Diseases)}.
\end{equation}
This setup reflects a single reasoning step where a query node is linked to multiple target nodes.

In 2-hop questions, the \textbf{query node} connects to the \textbf{target nodes} via an intermediate \textbf{bridge node}, forming a two-step reasoning chain. For example:
\begin{align}
    \text{Query (Phenotype)} &\xrightarrow{\text{side\_effects\_of}} \text{Bridge (Drug)} \nonumber \\
    &\xrightarrow{\text{treats}} \text{Target (Diseases)}.
\end{align}
Here, the bridge node (e.g., drug), used to query for 1-hop questions, serves as the intermediate entity linking the query and target.

\paragraph{Answer Definition.}
The \textbf{target nodes} are the final answers to the query. For 1-hop reasoning, this corresponds to all nodes directly connected to the query node. For 2-hop reasoning, the answers are all nodes that are accessible through the graph traversal via the bridge node, requiring models to infer both the intermediate (bridge) and final (target) nodes.

\subsection{Dataset Construction Pipeline}
The dataset is constructed using the following systematic process:
\begin{enumerate}
    \item \textbf{Entity Sampling:} Nodes representing drugs, diseases, proteins, and phenotype entities are extracted from PrimeKG.
    \item \textbf{2-Hop Path Definition:} For 2-hop questions, valid paths are constructed by combining two connected edges, ensuring the \textbf{query-bridge-target} structure follows the one-to-many-to-many relationship:
    \begin{equation}
        \text{Query} \xrightarrow{\text{Relation}_1} \text{Bridge} \xrightarrow{\text{Relation}_2} \text{Target}.
    \end{equation}
    \item \textbf{1-Hop Relationship Extraction:} For 1-hop questions, all relationships connecting query nodes (e.g., drugs) to their target nodes (e.g., diseases) are extracted. To maintain consistency with 2-hop questions, 1-hop relationships without a corresponding 2-hop path are excluded.
    \item \textbf{Answer Extraction:} For each question, all target nodes accessible through the graph traversal are extracted as answers. This ensures that the multi-answer nature of the dataset is preserved.
\end{enumerate}

\subsection{1-Hop and 2-Hop Questions}
\paragraph{1-Hop Questions.}
For a 1-hop question, the model is required to directly link the query node to the target node, which is populated using a template \textit{"Name a \textbf{Type}(\{Query\}) that is \textbf{Label}(\{Relationship\}) by \{Query\}"}. For example,
\begin{equation}
    \text{``Name a disease that is treated by Drug } Dr\text{?''}
\end{equation}
with the answer set defined as:
\begin{equation}
    A = \{D_1, D_2, \dots, D_n\},
\end{equation}
where $D_i$ represents disease linked to the query drug $Dr$.

\paragraph{2-Hop Questions.}
For a 2-hop question, the model must infer both the bridge node and the target node. The question template is \textit{"Name a \textbf{Type}(\{Query\}) that is \textbf{Label}(\{Relationship2\}) by \{Bridge\} that has a \textbf{Label}(\{Relationship1\}) \{Query\}"}. An example query is:
\begin{align}
    \text{``Name a disease that is treated by a drug}  \nonumber \\
    \text{that has a side effect }  S\text{?''}
\end{align}
The model needs to traverse the graph through an intermediate \textbf{bridge node} (drug) before reaching the final \textbf{target node} (disease):
\begin{equation}
    A = \{D_1, D_2, \dots, D_n\},
\end{equation}
where $D_i$ represents disease linked to the phenotype that is a side effect of drug $Dr$.

\subsection{Dataset Statistics}

\begin{table}[htbp]
\centering
\begin{tabular}{p{4.5cm}r}
\hline
\textbf{Relation (Query:Target)} & \textbf{Count} \\
\hline
\texttt{Protein:Disease}        & 731  \\
\texttt{Protein:Drug}           & 589  \\
\texttt{Disease:Drug}           & 297  \\
\texttt{Drug:Phenotype}         & 248  \\
\texttt{Drug:Disease}           & 234  \\
\texttt{Drug:Protein}           & 165  \\
\texttt{Disease:Protein}        & 113  \\
\texttt{Disease:Phenotype}      & 79   \\
\texttt{Phenotype:Drug}         & 33   \\
\texttt{Phenotype:Disease}      & 5    \\
\hline
\end{tabular}
\caption{Distribution of 1-hop relations in BioHopR.}
\label{tab:1hop_relations}
\end{table}

The \textbf{BioHopR} dataset consists of \textbf{2,494} unique 1-hop questions and \textbf{7,633} unique 2-hop questions, resulting in a total of \textbf{279,738} answers. On average, each question is associated with \textbf{36.65} answers, reflecting the dataset's complexity and the many-to-many relationships inherent in biomedical knowledge. The dataset includes 10 distinct 1-hop relation types and 12 2-hop relation types, and the breakdown of the number of questions for each relation type is summarized in Tables \ref{tab:1hop_relations} and \ref{tab:2hop_relations}.

\begin{table}[htbp]
\centering
\begin{tabular}{p{5.8cm}r}
\hline
\textbf{Relation (Query:Bridge:Target)} & \textbf{Count} \\
\hline
\texttt{Drug:Protein:Disease}        & 3029 \\
\texttt{Disease:Drug:Phenotype}      & 949  \\
\texttt{Disease:Protein:Drug}        & 899  \\
\texttt{Protein:Disease:Drug}        & 577  \\
\texttt{Phenotype:Disease:Drug}      & 546  \\
\texttt{Protein:Drug:Disease}        & 462  \\
\texttt{Disease:Drug:Protein}        & 381  \\
\texttt{Drug:Disease:Protein}        & 321  \\
\texttt{Phenotype:Drug:Disease}      & 215  \\
\texttt{Drug:Disease:Phenotype}      & 213  \\
\texttt{Disease:Phenotype:Drug}      & 36   \\
\texttt{Drug:Phenotype:Disease}      & 5    \\
\hline
\end{tabular}
\caption{Distribution of 2-hop relations in BioHopR.}
\label{tab:2hop_relations}
\end{table}

The restriction to one-to-many-to-many relationships ensures that the dataset mirrors real-world biomedical reasoning scenarios, where single entities often relate to multiple downstream entities. This design makes the dataset uniquely suited for evaluating large language models (LLMs) on tasks requiring multi-step reasoning and comprehensive answer generation.

\subsection{Qualitative Analysis}
To better understand the models' reasoning capabilities, we conducted a qualitative analysis on the questions about Type II Diabetes, as it is one of the widely studied diseases \cite{skyler2017differentiation}.

\subsection{Reasoning Benchmark}
BioHopR presents significant reasoning challenges:
\begin{itemize}[noitemsep, topsep=0pt]
    \item Models must implicitly identify intermediate bridge nodes in 2-hop questions while ensuring the correctness of the final answers.
    % \item Generating comprehensive multi-answer outputs is critical, as each question may have numerous valid target nodes.
    \item The many-to-many nature of biomedical relationships requires models to handle diverse answer sets while preserving reasoning consistency.
\end{itemize}

\section{Experiments}
\label{sec:experiments}

We evaluate a range of LLMs on the BioHopR benchmark to assess their ability to reason over one-to-many-to-many relationships. The evaluation focuses on both single-answer and multi-answer reasoning for 1-hop and 2-hop questions, highlighting the challenges posed by multi-step reasoning and comprehensive answer generation.

\subsection{Experimental Setup}

\paragraph{Models Evaluated.}
We consider a diverse set of LLMs, categorized into general-purpose proprietary, reasoning proprietary, medical-specific, and open-source models, as detailed in Table \ref{tab:models}. General-purpose models include GPT4O and smaller variants such as GPT4O-mini \cite{hurst2024gpt}. We also added O3-mini as it was most recent cost-effective reasoning proprietary model \cite{openai_o3_mini}. We also evaluate open-source Llama models (Llama3.1 and Llama3.3) with varying parameter scales (8B and 70B) \cite{dubey2024llama}. We selected medical-specific models that are based on the baseline Llama3.1 architectures: UltraMedical-8B, HuatuoGPT-o1-8B, and HuatuoGPT-o1-70B \cite{zhang2024ultramedical,chen2024huatuogpt}. HuatuoGPT-o1 models are trained for medical complex reasoning for medical problems.

\begin{table}[htbp]
\centering
\begin{tabular}{p{3.2cm}p{3.2cm}}
\hline
\textbf{Model Name} & \textbf{Domain} \\
\hline
GPT4O & General \\
GPT4O-mini & General \\
O3-mini & Reasoning  \\ \hline
Llama3.1 8B & General \\
Llama3.1 70B & General  \\
Llama3.3 70B & General  \\
UltraMedical-8B & Medical  \\
HuatuoGPT-o1-8B & Medical Reasoning \\
HuatuoGPT-o1-70B & Medical Reasoning \\
\hline
\end{tabular}
\caption{Models evaluated in the experiments.}
\label{tab:models}
\end{table}

\begin{table*}[t]
\centering
\begin{tabular}{lcccc}
\hline
\textbf{Model} & \textbf{Prec\_HOP1 (\%)} & \textbf{Prec\_HOP2 (\%)} & \textbf{BOTH\_COR (\%)} & \textbf{BOTH\_WR (\%) ↓} \\
\hline
Llama-3.1-8B          & 0.12 & 0.05 & 0.00 & 99.76 \\
HuatuoGPT-o1-70B         & 0.16 & 0.00 & 0.00 & 99.93 \\
HuatuoGPT-o1-8B          & 0.20 & 0.04 & 0.00 & 99.54 \\
UltraMedical-8B       & 13.75 & 5.21 & 2.28 & 82.33 \\
Llama-3.3-70B         & 25.58 & 9.58 & 4.94 & 68.33 \\
Llama-3.1-70B         & 26.38 & 9.47 & 4.93 & 65.64 \\
GPT4O-mini            & 28.11 & 14.57 & 6.54 & 64.69 \\
GPT4O                 & 32.88 & 14.57 & 7.86 & 57.96 \\
O3-mini               & \textbf{37.93} & \textbf{14.57} & \textbf{8.93} & \textbf{52.14} \\
\hline
\end{tabular}
\caption{Performance metrics (in percentages) for various models. Prec\_HOP1 and Prec\_HOP2 represent the precision on 1-hop and 2-hop tasks, respectively. BOTH\_COR indicates cases where both hops are correct, and BOTH\_WR indicates cases where both hops are incorrect (the lower the better).}
\label{tab:model_performance_compact}
\end{table*}

\subsection{Evaluation}
The proprietary GPT models (GPT4O, GPT4O-mini, and O3-mini) were accessed using OpenAI’s API\footnote{\url{https://platform.openai.com/docs/models}}. For open-source models, we used four A100 GPUs with 80GB memory per GPU for 70B parameter models and one A6000 GPU for 8B parameter models. The evaluation was conducted in a zero-shot setting, with a batch size of 1 and a temperature set to 0, except O3-mini model which does not support temperature parameter, to ensure deterministic responses. The evaluation code for open-source models were implemented using the HuggingFace Transformers library \cite{wolf2019huggingface}.

\subsection{Evaluation Metrics}

\paragraph{Embedding-Based Precision.}
The precision ($\text{Prec}$) is computed using the cosine similarity between the predicted response and the ground truth answer list, leveraging BioLORD-2023-C embeddings \cite{remy2023biolord}. Let $p$ denote the embedding of the predicted response and $\{a_1, a_2, \dots, a_n\}$ denote the embeddings of the ground truth answers. The cosine similarity for a prediction $p$ and an answer $a_i$ is defined as:
\begin{equation}
    \text{cos}(p, a_i) = \frac{p \cdot a_i}{\|p\| \|a_i\|}.
\end{equation}

If the maximum cosine similarity across all ground truth answers satisfies:
\begin{equation}
    \max_{i \in \{1, \dots, n\}} \text{cos}(p, a_i) > \tau,
\end{equation}
then the prediction is considered a true positive. 
The precision ($\text{Prec}$) is then calculated as:
\begin{equation}
    \text{Prec} = \frac{|\text{True Positives}|}{|\text{Predicted Responses}|}
\end{equation}.

We use $\tau = 0.9$ for BioLORD-2023-C embeddings after a grid search of threshold values from 0.5 to 0.9, which led an optimal setting with 0.9. Please refer to more details in the Appendix Figure \ref{fig:cosine_threshold}. This threshold prioritizes precision, ensuring that only highly confident predictions are accepted as correct. By setting a high threshold, we align with the strict requirements of biomedical applications, minimizing false positives while maintaining robust handling of biomedical definition-level similarity and ambiguous synonyms.

\section{Results and Discussion}
\label{sec:results}

\subsection{Proprietary Models Demonstrate Robust Multi-Hop Reasoning}

Proprietary models (GPT4O, GPT4O-mini, and O3-mini) demonstrate consistently strong performance across all metrics. For 1-hop tasks (\textbf{Prec\_HOP1}), O3-mini achieves the highest precision (37.93\%), followed by GPT4O (32.88\%) and GPT4O-mini (28.11\%). Interestingly, all proprietary models achieve identical performance on 2-hop tasks (\textbf{Prec\_HOP2: 14.57\%}), suggesting a possible shared capabilities for implicit reasoning or complex reasoning.

These results reflect the impact of the reasoning step before answering. O3-mini’s higher \textbf{Prec\_HOP1} indicates the reasoning capability of the model allowed it to reason well on single-step queries.

\subsection{Open-Source Biomedical Models Face Significant Challenges}

Open-source biomedical models struggle to match the performance of proprietary models, particularly on multi-hop tasks. HuatuoGPT-o1 models perform the worst, achieving near-zero precision for both 1-hop (\textbf{Prec\_HOP1: 0.20\%} for HuatuoGPT-o1-8B) and 2-hop (\textbf{Prec\_HOP2: 0.00\%} for HuatuoGPT-o1-70B). In contrast, UltraMedical-8B performs better (\textbf{Prec\_HOP1: 13.75\%}, \textbf{Prec\_HOP2: 5.21\%}).

These results suggest that although HuatuoGPT-1 was trained for medical complex reasoning, it's generalizability is far less than UltraMedical. The reasoning demands of BioHopR is far different from medical license examination based QA datasets such as MedQA, which HuatuoGPT-o1 used for training. Still UltraMedical-8B’s performance, when compared to a larger general domain open-source models such as Llama3.1-70B and Llama3.3-70B, is far behind, suggesting persistent challenges in resolving bridge nodes for multi-hop queries.

\paragraph{Error Patterns.} 
The \textbf{BOTH\_WR} metric reveals systemic challenges in multi-hop reasoning for all models. Open-source models like HuatuoGPT-o1-70B exhibit the highest \textbf{BOTH\_WR} rates (\textgreater 99\%), reflecting widespread failure in both reasoning hops. Proprietary models demonstrate significantly lower failure rates, with O3-mini achieving the best performance (\textbf{BOTH\_WR: 52.14\%}). However, even the best-performing models show substantial error rates in both hops, indicating that multi-step inference remains a bottleneck.

\subsection{Multi-Hop Reasoning Remains a Bottleneck}

Across all models, performance declines sharply from 1-hop to 2-hop tasks. For example, GPT4O’s precision drops from \textbf{Prec\_HOP1: 32.88\%} to \textbf{Prec\_HOP2: 14.57\%}, while open-source models like Llama-3.1-8B exhibit near-complete failure (\textbf{Prec\_HOP2: 0.05\%}).

This decline highlights the inherent complexity of multi-hop reasoning. Resolving 2-hop queries requires implicit inference of intermediate entities (e.g., bridge nodes) and alignment of reasoning chains across multiple steps.

\subsection{Qualitative Analysis - Case Studies}
Our qualitative analysis on various diseases, including Type II Diabetes and Schizophrenia aligns well with the evaluation result in Table \ref{tab:model_performance_compact}. We highlight the diabetes-related questions in Figure \ref{fig:diabetes_analysis}.

\begin{figure}[h]
% \scriptsize
\footnotesize
\centering
\begin{tabular}{p{1.8cm}|p{2.2cm}|p{2.2cm}}
\toprule
\multicolumn{3}{c}{\textbf{Questions}} \\ \midrule
 \multicolumn{3}{l}{\textbf{Hop1}: \textit{"...a side effect of drug Troglitazone."}} \\ 
 \multicolumn{3}{l}{\textbf{Hop2}: \textit{"...a side effect of a drug ... treat type 2 diabetes."}} \\ \midrule
\textbf{Model} & \textbf{Hop1 Prediction} & \textbf{Hop2 Prediction} \\ \midrule

\textbf{HuatuoGPT-o1-70B} & 
\textcolor{red}{\textit{"Alright, let's think about Troglitazone..."}} & 
\textcolor{red}{\textit{"Alright, let's think about this..."}} \\ \midrule

\textbf{HuatuoGPT-o1-8B} & 
\textcolor{orange}{\textit{"Hepatotoxicity"}} & 
\textcolor{blue}{\textit{"Hypoglycemia"}} \\ \midrule

\textbf{UltraMedical-8B} & 
\textcolor{orange}{\textit{"Hepatotoxicity"}} & 
\textit{\textcolor{blue}{"Lactic acidosis}, \textcolor{blue}{Hypoglycemia}, \textcolor{red}{Hyperkalemia"}} \\ \midrule

\textbf{GPT4O} & 
\textcolor{orange}{\textit{"Hepatotoxicity"}} & 
 \textcolor{red}{\textit{"Weight gain"}}  \\ \midrule

\textbf{O3-mini} & 
\textcolor{orange}{\textit{"Hepatotoxicity"}} & 
 \textcolor{red}{\textit{"Weight gain"}}  \\ \bottomrule
\end{tabular}
\caption{Qualitative analysis of model responses to diabetes-related questions. Red-colored text shows the wrong answer. Orange-colored text shows the answer that is not in the answer list, but is plausible. Blue-colored text shows the correct answer.}
\label{fig:diabetes_analysis}
\end{figure}

\begin{table*}[htbp]
\centering
\begin{tabular}{p{4.5cm}cccccc}
\hline
\multirow{3}{*}{\textbf{Relation Type}} & \multicolumn{2}{c}{\textbf{1-Hop}} & \multicolumn{2}{c}{\textbf{2-Hop}} \\
 & \textbf{GPT4O} & \textbf{GPT4O-mini} & \textbf{GPT4O} & \textbf{GPT4O-mini } \\
 & (Single/Multi) & (Single/Multi) & (Single/Multi) & (Single/Multi)\\
\hline
\multicolumn{5}{c}{\textbf{Same Query and Bridge}} \\ \hline
\texttt{Disease:Drug:Phenotype}      & 47.47 / 36.49 & 43.77 / 26.56        & 25.08 / 4.71  & 25.08 / 4.01  \\
\texttt{Disease:Drug:Protein}        & 47.47 / 36.49 & 43.77 / 26.56        & 3.67 / 1.13   & 3.67 / 1.29   \\
\texttt{Drug:Disease:Phenotype}      & 55.13 / 13.67 & 50.85 / 14.75        & 22.07 / 6.77  & 22.07 / 7.12 \\
\texttt{Drug:Disease:Protein}        & 55.13 / 13.67 & 50.85 / 14.75        & 4.67 / 0.27   & 4.67 / 0.43   \\  \hline
\multicolumn{5}{c}{\textbf{Same Query and Target}} \\  \hline
\texttt{Disease:Phenotype:Drug}      & 20.25 / 8.48  & 22.78 / 7.14        & 16.67 / 2.38  & 16.67 / 1.50  \\
\texttt{Disease:Protein:Drug}        & 35.40 / 7.8  & 26.55 / 2.93        & 8.12 / 4.60   & 8.12 / 2.81   \\
\texttt{Drug:Phenotype:Disease}      & 23.39 / 4.94  & 31.05 / 3.70        & 0.00 / 0.00   & 0.00 / 1.01   \\
\texttt{Drug:Protein:Disease}        & 20.61 / 5.77  & 20.00 / \textbf{44.00}         & 20.14 / 1.84  & 20.14 / 2.82  \\  \hline
\multicolumn{5}{c}{\textbf{Others}} \\  \hline
\texttt{Phenotype:Disease:Drug}      & 0.00 / \textbf{8.08}   & 0.00 / \textbf{3.03}         & 14.47 / \textbf{17.24}  & 14.47 / 6.86  \\
\texttt{Phenotype:Drug:Disease}      & 15.15 / 10.00  & 24.24 / 2.43         & 3.72 / 3.26   & 3.72 / 2.40   \\
\texttt{Protein:Disease:Drug}        & 35.29 / 7.20 & 27.50 / 5.26        & 2.95 / 1.89   & 2.95 / 0.48  \\
\texttt{Protein:Drug:Disease}        & 23.60 / 20.00  & 14.43 / 9.35         & 1.08 / \textbf{1.40}   & 1.08 / \textbf{1.40}   \\

\hline
\textbf{Overall}                     & \textbf{32.88} / \textbf{8.09}  & 28.11 / 6.11  & \textbf{14.57} / \textbf{3.46}  & 14.57 / 3.18  \\
\hline
\end{tabular}
\caption{Comparison of Single-Answer prompting and Multi-Answer prompting for GPT4O and GPT4O-mini across 1-hop and 2-hop relation types. Precision (Prec) is reported.}
\label{tab:combined_comparison}
\end{table*}

Diabetes-related questions for drug Troglitazone, which has 202 side effects listed from PrimeKG, highlighted mixed performance among models. For instance, HuatuoGPT-o1-8B correctly predicted answers but diverged from the task constraints by elaborating on its reasoning instead of adhering to the prompt. Similarly, UltraMedical produced multiple answers when a single response was requested, with only some of the predictions being correct. In contrast, proprietary models such as GPT-4 reliably adhered to prompted task, consistently including relevant responses such as hepatotoxicity, even if these were not explicitly part of the predefined answer set. This behavior suggests that proprietary models may apply broader medical reasoning compared to open-source models. Proprietary models generally outperform open-source models in both task adherence and reasoning accuracy.

\subsection{Ablation Study: Prompting Strategy}

\paragraph{Prompting Setup.}
The dataset can also support multi-answer prompting, making two prompting strategies designed to evaluate different aspects of model reasoning:
\begin{itemize}[noitemsep, topsep=0pt]
    \item \textbf{Single-Answer Prompting:} The model is prompted to provide one correct answer (e.g., \textit{“Name a gene associated with Disease X.”}). This evaluates the model’s ability to identify the most probable answer using implicit reasoning.
    \item \textbf{Multi-Answer Prompting:} The model is prompted to provide all correct answers (e.g., \textit{“Name all genes associated with Disease X.”}). This evaluates the model’s ability to generate exhaustive, comprehensive outputs, which is inherently more challenging.
\end{itemize}

While both strategies are valuable for understanding model performance, \textbf{Multi-Answer Prompting} poses significant challenges. On average, each question in the dataset has 36.65 correct answers, making it computationally expensive and cognitively demanding for large language models to generate a complete answer set.

\subsubsection{Evaluation Metric for Multi-Answer}

\paragraph{Precision for Multi-Answer Prompting.}
For \textbf{Multi-Answer Prompting}, precision is computed using cosine similarity-based matching. Let $P = \{p_1, p_2, \dots, p_m\}$ denote the embeddings of the predicted responses and $A = \{a_1, a_2, \dots, a_n\}$ denote the embeddings of the ground truth answers. A predicted response $p_j$ is considered a true positive if:
\begin{equation}
    \max_{i \in \{1, \dots, n\}} \text{cos}(p_j, a_i) > \tau,
\end{equation}
where we set $\tau = 0.9$ for high-confidence matches. Then the precision is computed in the same way as the \textbf{Single-Answer Prompting}.

\subsubsection{Analysis and Results.}

To analyze the feasibility of Multi-Answer Prompting, we conducted an ablation study using GPT4O and GPT4O-mini, the proprietary models. Table \ref{tab:combined_comparison} presents the performance metrics for Single-Answer and Multi-Answer prompting across 1-hop and 2-hop tasks.

\paragraph{Single-Answer Prompting Outperforms Multi-Answer Prompting:} GPT4O achieves an average Prec of 32.88\% in 1-hop tasks, significantly higher than its Multi-Answer score of 8.09\%. The gap is even more pronounced in 2-hop tasks, where GPT4O achieves 14.57\% Prec compared to just 3.46\%. A similar pattern holds for GPT4O-mini, showing 28.11\% vs. 6.11\% in 1-hop and 14.57\% vs. 3.18\% in 2-hop respectively. Relations with abstract or less structured targets (e.g., \texttt{Disease:Drug:Protein} and \texttt{Drug:Phenotype:Disease}) exhibit particularly poor precision scores under Multi-Answer prompting, with both models achieving below 2\% in 2-hop tasks. These results highlight the difficulty of generating comprehensive answer sets, especially for complex reasoning paths.

\paragraph{Cases Where Multi-Answer Prompting Performs Competitively.}
Despite the overall trend, there are a few relation types where Multi-Answer prompting yields surprisingly competitive or even stronger results. For example, in the 2-hop relation \texttt{Phenotype:Disease:Drug}, GPT4O achieves 17.24\% precision under Multi-Answer prompting—exceeding its Single-Answer precision of 14.47\%. Similarly, in 1-hop tasks such as \texttt{Phenotype:Disease} and \texttt{Protein:Drug}, GPT4O-mini achieves 44.00\% with Multi-Answer prompting compared to 20.00\% under Single-Answer prompting. These cases suggest that for certain structured relations with high overlap between bridge and target entities, models may benefit from listing multiple answers rather than specifying a specific answer.

Based on these findings, we restricted our evaluation of all other models to Single-Answer Prompting. This decision is motivated by higher robustness and computational overhead of multi-answer prompting. Also in many real-world scenarios, users typically seek the most probable or relevant answer, aligning more closely with Single-Answer prompting.

While Multi-Answer prompting offers valuable insights into a model’s ability to generate exhaustive outputs, it remains a challenging evaluation paradigm. Future work could focus on improving model training and prompting strategies to better support comprehensive answer generation.

\section{Conclusion} \label{sec:conclusion}

We introduced \textbf{BioHopR}, a benchmark for evaluating multi-hop, multi-answer reasoning in the biomedical domain. Built on the PrimeKG knowledge graph, BioHopR captures the complexity of real-world biomedical queries through \textit{one-to-many} and \textit{many-to-many} relationships, rigorously assessing reasoning over 1-hop and 2-hop tasks.

Evaluation results highlight that O3-mini, a proprietary model with a reasoning step, outperforms open-source models including biomedical models like HuatuoGPT-o1. Across all models, the performance drop from 1-hop to 2-hop tasks underscores the difficulty of aligning intermediate reasoning steps, especially in bridging entities.

By addressing the lack of benchmarks for multi-hop reasoning in biomedical domain, BioHopR sets a new standard for evaluating reasoning capabilities and provides a critical step toward more robust and interpretable LLMs for biomedical research and real-world applications. Future directions include expanding the dataset to other knowledge sources and domains, such as chemistry.

\section*{Limitation}
\label{sec:limitation}
While BioHopR provides a rigorous benchmark for evaluating multi-hop reasoning in the biomedical domain, several limitations exist. BioHopR is currently focused on 4 major entities only: Protein, Phenotype, Drug, Disease. Also, it relies exclusively on a single knowledge graph, PrimeKG, which, while comprehensive, may not fully capture the diversity of biomedical knowledge or its real-world dynamics. This lack of diversity could bias model evaluation toward the structure and content of 4 major node types and PrimeKG, potentially under-representing a model's ability to generalize to other knowledge and sources. While human evaluation was not the primary focus of this work, future efforts could include more extensive and diverse human evaluations to validate model-generated outputs.

\section*{Broader Impacts and Ethics Statement}
\label{sec:ethics}
Our work raises no major ethical concerns. All evaluations and experiments were conducted strictly for research purposes. 

We will release BioHopR. License and copyright information, along with Terms of Use, will be made available upon release of the dataset and associated materials. While BioHopR facilitates advancements in biomedical reasoning tasks, it is not designed for use in real-world clinical applications. Consequently, models evaluated or trained on BioHopR should not be used for clinical decision-making without rigorous validation and regulatory approvals.

This restriction aims to mitigate potential risks associated with incorrect reasoning or hallucinated outputs, which could lead to harmful clinical outcomes. Additionally, while BioHopR supports research into biomedical reasoning, it is critical that researchers use the benchmark responsibly, with appropriate safeguards in place to ensure the ethical use of derived insights and outputs.

% Bibliography entries for the entire Anthology, followed by custom entries
%\bibliography{anthology,custom}
% Custom bibliography entries only
\clearpage
\bibliography{acl_latex_v2}

\clearpage
\appendix
\section*{Appendix}
\label{sec:appendix}

\begin{figure*}[bp]
    \centering
    \includegraphics[width=1.0\linewidth]{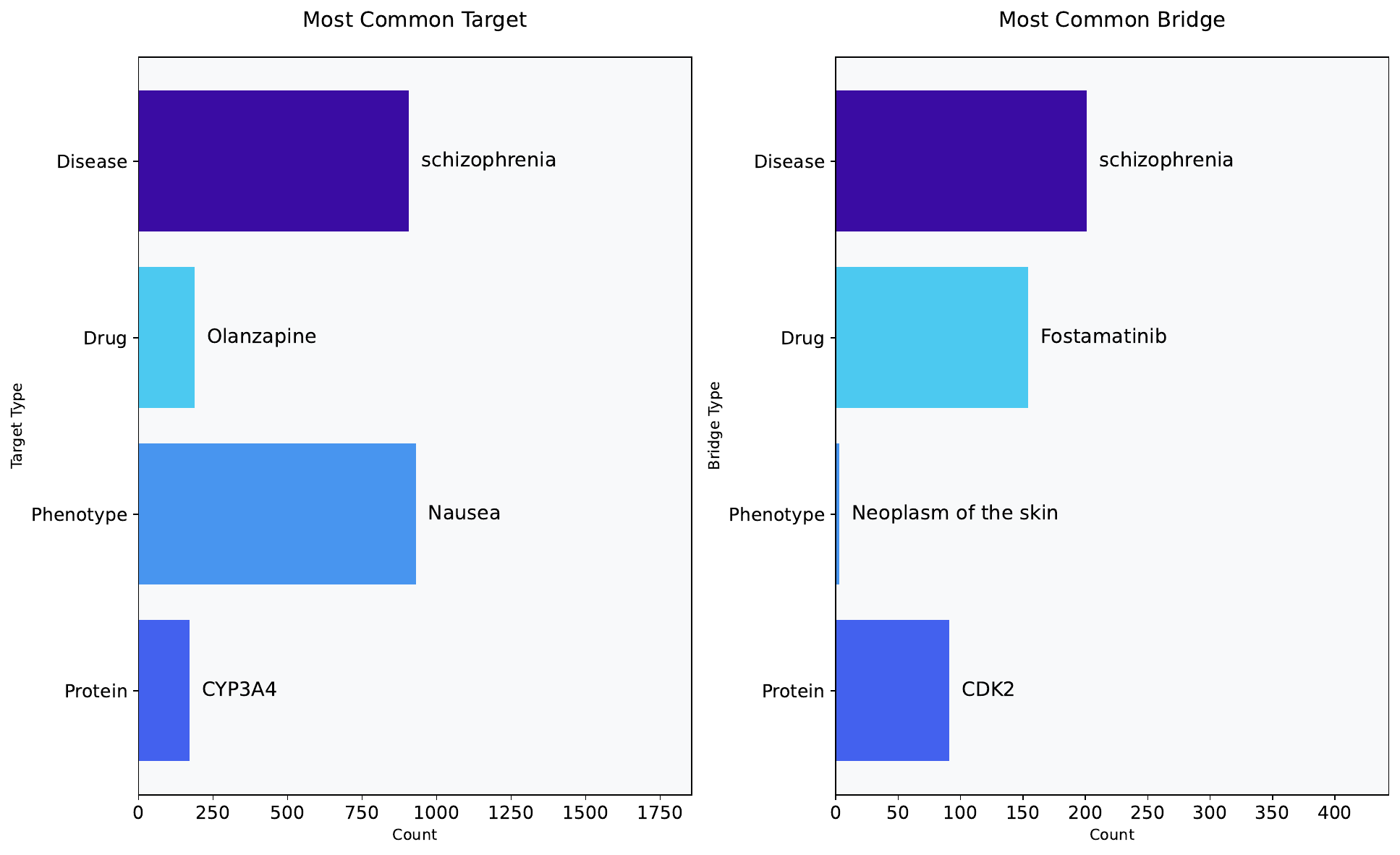}
    \caption{Common target and bridge entities for each node type in BioHopR.}
    \label{fig:comm_target_bridge}
\end{figure*}

The Figure \ref{fig:comm_target_bridge} illustrates the frequency distribution of target and bridge entities within the BioHopR dataset, highlighting key patterns. The left panel demonstrates the prevalence of proteins (e.g., CYP3A4), phenotypes (e.g., Nausea), drugs (e.g., Olanzapine), and diseases (e.g., Schizophrenia) as target nodes in multi-hop queries. Meanwhile, the right panel showcases the distribution of bridge entities, which frequently include proteins (e.g., CDK2), phenotypes (e.g., Neoplasm of the skin), drugs (e.g., Fostamatinib), and diseases. These patterns reflect the diversity and real-world complexity of biomedical entities, emphasizing the challenges of reasoning over structured knowledge graphs for multi-hop queries.

\begin{figure*}[htbp]
\centering
\includegraphics[width=1.0\linewidth]{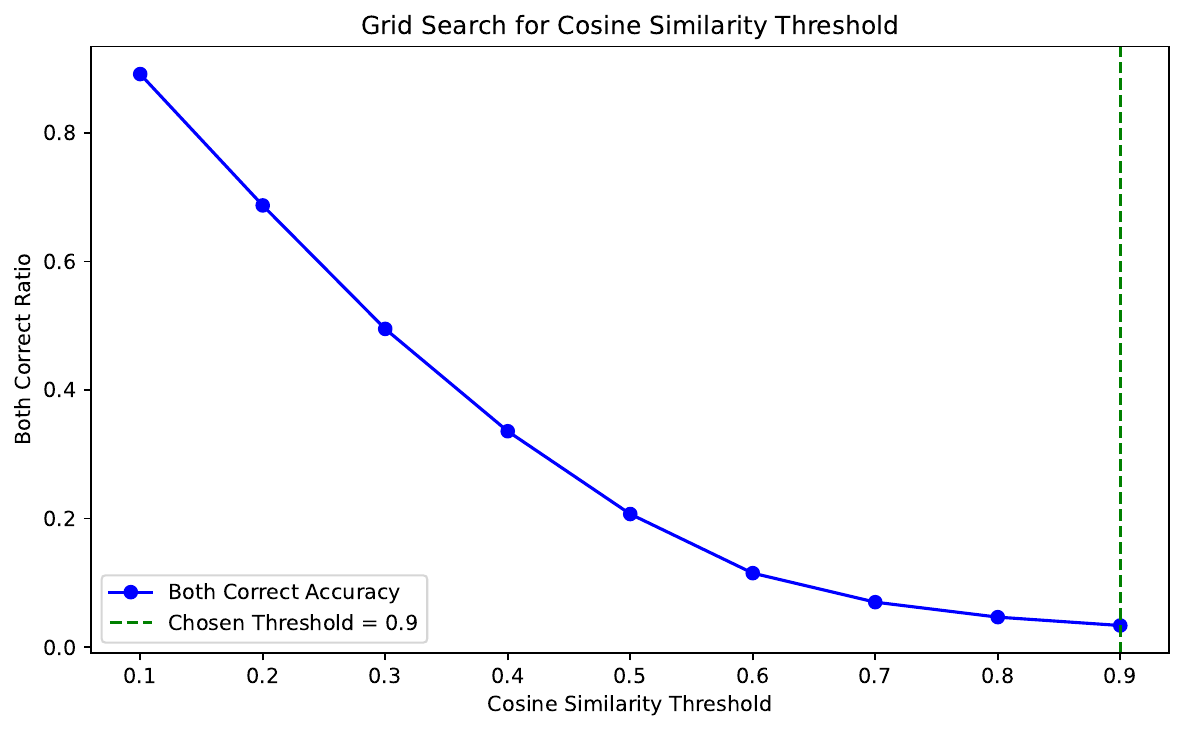}
\caption{Grid search results showing the relationship between cosine similarity threshold and accuracy for "Both Correct" predictions. The chosen threshold of 0.9 is marked, reflecting the strict precision requirements in the biomedical domain.}
\label{fig:cosine_threshold}
\end{figure*}

Figure \ref{fig:cosine_threshold} illustrates the results of a grid search for determining the optimal cosine similarity threshold for BioLORD-2023-C embeddings. The x-axis represents the threshold values, ranging from 0.1 to 0.9, while the y-axis shows the accuracy for "Both Correct" predictions. A sharp decline in accuracy is observed as the threshold increases, with accuracy plateauing beyond 0.8. The chosen threshold of 0.9 ensures high precision by accepting only highly confident predictions, aligning with the strict requirements of biomedical reasoning tasks.

\section{Detailed Qualitative Analysis}
We further included other diseases: Vitamin A Deficiency, Lung Cancer, Alzheimer’s Disease, Schizophrenia. We selected these medical conditions because they represent a range of domains within the biomedical field, which include nutritional deficiencies, metabolic disorders, chronic diseases and neurodegenerative conditions. This selection allows for a more comprehensive assessment of the models' ability to reason across different medical contexts and complexities. Additionally, for conditions such as Type II diabetes and Vitamin A deficiency, the answers may seem quite straightforward, making them useful for assessing whether the models can correctly identify and reason over well-established medical knowledge. Whereas, for complex conditions such as Lung Cancer and Alzheimer's Disease, we can evaluate the models ability to reason through more intricate, multi-factorial diseases.

Our qualitative analysis showed several key findings regarding the models' reasoning capabilities across different diseases. Interestingly, none of the models generated questions for Alzheimer's Disease, which was unexpected given its significant global impact and strong presence in the Knowledge Graph. In contrast, the models seemed to reason well over diabetes-related questions, although it would often provide multiple correct answers, even when prompted for a single response. This could suggest an alignment with well-established medical knowledge in this domain. For cancer-related questions, the models tended to select the most straightforward and common answers, though the Knowledge Graph contained a broader mix of more complex phenotypes. This seems to indicate a preference for simplicity in model-generated reasoning, potentially overlooking more nuanced aspects of the disease.

When comparing open-source models against proprietary models, the qualitiative analysis shows that proprietary models generally performed better than open-source models in terms of providing structured and direct responses. proprietary models demonstrated a better adherence to the prompt constraints, whilst the open source models seem to show more explanatory or multi-component answers. For example, the HuatuoGPT-70B open source model, consistently responded with "thinking" before elaborating on its reasoning instead of directly providing a single answer for both 1-hop and 2 hop prediction as prompted. This suggests that the model prioritises explaining its reasoning than strictly following the prompt's format. In constrast, however, proprietary models such as GPT-4 more reliably adhered to the prompt constraints,. When prompted to give a single answer for one hop and two hop questions, GPT-4 consistently did so, and this was present across the closed-source GPT family, suggesting that these proprietary models may be better optimised for tasks requring direct and efficient responses. Among the open source models tested, LLaMA Ultra Medical, a medical open source LLM, tended to provide multiple answers when prompted for one single answer, and of those multiple answers, apart from Type II diabetes, most answers were incorrect.  

Taking an look into responses related to diabetes, responses were quite mixed. For instance, HuatuoGPT-01-8B performed outside the constraints of the task, correctly predicting the answer before proceeding to provide its reasoning. On the other hand,  LLaMA 8B Instruct struggled with both Hop 2 and Hop 1 predictions, failing to generate the correct responses. Similarly, LLaMA Ultra Medical did not fully adhere to the prompt’s instructions — when asked to provide a single answer for Hop 2, it instead generated a list of multiple possible answers. While the listed responses were correct, this deviation indicates a challenge in following explicit task constraints. Moreover, for Hop 1, the model's response was incorrect, further highlighting inconsistencies in its performance. Interestingly, GPT-4 did not correctly predict the Hop 2 or Hop 1 answers in a strict sense. However, the model consistently included hepatotoxicity as a response—a condition that, while not explicitly listed among the correct answers, is still a relevant and justifiable finding. This pattern was observed across the GPT model family, suggesting that these models might apply broader medical reasoning even when their direct predictions do not align with predefined correct answers.

Schizophrenia, as seen in our figure, appeared frequently in the data. For GPT-4, in one-hop predictions, the model frequently guessed Clozapine as a treatment for schizophrenia. While this answer is medically correct, it was not explicitly part of the datasets predefined answer set. This suggests that the model is leveraging broader clinical knowledge rather than strictly adhering to the dataset's constraints. This trend was also consistent across the other GPT-family models, GPT-4o mini. However, for the o3 models, the one-hop predictions were correct and within our predefined list of answers. 

\end{document}